\pdfoutput=1
\documentclass[runningheads]{llncs}
\usepackage{graphicx}
\usepackage{tikz}
\usepackage{comment}
\usepackage{amsmath,amssymb} 
\usepackage{color}
\usepackage{algorithm}
\usepackage[noend]{algpseudocode}
\usepackage{subcaption}
\usepackage{siunitx}
\usepackage{stackengine}
\begin{document}
\pagestyle{headings}
\mainmatter
\title{Towards Robust Low Light Image Enhancement}
\titlerunning{Towards Robust Low Light Image Enhancement}
%
\author{Sara Aghajanzadeh\inst{1}
\and
David Forsyth\inst{1} 
}
\authorrunning{S. Aghajanzadeh and D. Forsyth}
%
\institute{University of Illinois Urbana Champaign, Champaign IL 61820, USA
\email{saraa5@illinois.edu}\\
\email{daf@illinois.edu}\\}

\maketitle

\begin{abstract}
Images captured in poor illumination suffer from degraded quality. We propose a low light image enhancement solution to produce visually pleasing normal light images. We use the abundant data available in the wild to train a strong model. We rely on a straightforward simulation of an imaging pipeline to generate usable dataset for training and testing. We demonstrate the generalization power of our approach using zero-shot cross-dataset transfer, i.e., we evaluate on datasets that were never seen during training. Our approach outperforms competing methods across a number of standard datasets. 
\keywords{Low light image enhancement, zero-shot cross-dataset transfer}
\end{abstract}

\section{Introduction}
\label{sec:introduction}

\begin{figure*}[!ht]
	\centering
	\begin{subfigure}{0.22\textwidth}
		\includegraphics[width=\textwidth]{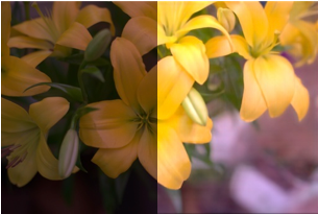}
	\end{subfigure}
	\begin{subfigure}{0.22\textwidth}
		\includegraphics[width=\textwidth]{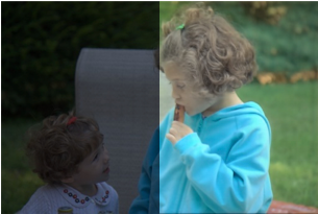}
	\end{subfigure}
	\begin{subfigure}{0.152\textwidth}
		\includegraphics[width=\textwidth]{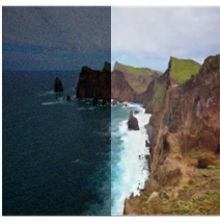}
	\end{subfigure}
	\begin{subfigure}{0.148\textwidth}
		\includegraphics[width=\textwidth]{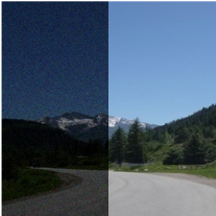}
	\end{subfigure}
	\begin{subfigure}{0.15\textwidth}
		\includegraphics[width=\textwidth]{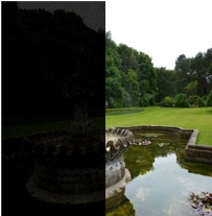}
	\end{subfigure}
	\begin{subfigure}{0.22\textwidth}
		\includegraphics[width=\textwidth]{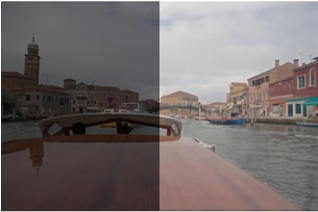}
	\end{subfigure}
	\begin{subfigure}{0.22\textwidth}
		\includegraphics[width=\textwidth]{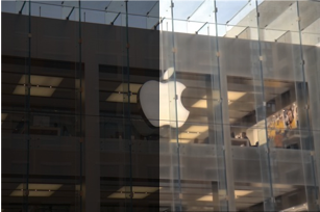}
	\end{subfigure}
	\begin{subfigure}{0.148\textwidth}
		\includegraphics[width=\textwidth]{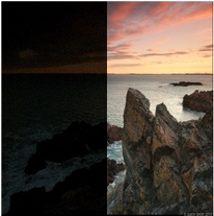}
	\end{subfigure}
	\begin{subfigure}{0.154\textwidth}
		\includegraphics[width=\textwidth]{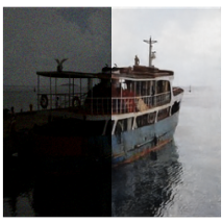}
	\end{subfigure}
	\begin{subfigure}{0.15\textwidth}
		\includegraphics[width=\textwidth]{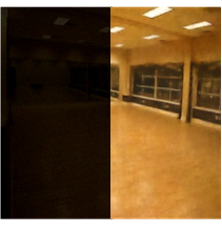}
	\end{subfigure}
	\caption{\label{fig:intro-fig} Our enhancement results (right) on low light images (left). Horizontally symmetric images are shown for fairness. First two columns show MIT-Adobe-FiveK images. Our model is neither trained nor fine-tuned on these images. Last three columns show our test-set. We can assess the model on different levels of darkness because of our stochastic low light images.}
\end{figure*}

This paper describes a method that takes
dark images found in the wild, makes them bright and gets the answer
right.  There are existing methods to do so, but each works best
on the test split of the dataset used in training.   In contrast,
we demonstrate {\it zero-shot cross-dataset
  transfer}~\cite{zero-shot-cross-dataset-transfer_2020} where we
supervise our model on diverse and representative data but evaluate on
a number of standard datasets that were never seen during training.  

Brightening a found dark image is hard for several reasons.  First,
the image is dark because it was taken in low light, so there are
color shifts caused by quantization and there is sensor noise.
Second, the camera response function used to obtain the image is
unknown, so correcting to RAW and then using a method adapted
to RAW is not possible.  Finally, there isn't much paired training
data.

We use a straightforward simulation of an imaging pipeline to
generate usable simulated training data.  We take arbitrary normal light
images and pass them through a randomly chosen inverse camera response function
(CRF) to linearize. We multiply the linearized image by a small randomly
chosen weight to make them dark, then simulate real low light images
by adding random shot and read noise. Finally, we pass the result
through another random CRF. Fig.~\ref{fig:method} illustrates our
procedure, which yields a very large simulated paired training dataset.

{\bf Contributions:}  We show that simulated paired data can be used
to train a straightforward but very effective brightener for found
images.  We exploit the analogy between low light image
enhancement and colorization problem~\cite{colorfu_2016l}
\cite{im2im_2018}, supervising the model with
images in LAB color space rather than RGB. 
Our approach outperforms the state of the art
quantitatively. Qualitative comparisons suggest strong improvements in
reconstruction accuracy (Fig.~\ref{fig:intro-fig} for some examples).

\section{Related Work}
\label{sec:related work}
Deep learning methods now dominate low light image enhancement.
A recent survey~\cite{survey_2021} gives a
comprehensive overview of existing works. Here, we summarize 
recent and representative methods which we use as baselines for
comparisons in Sec.~\ref{sec:experiment}, categorized by learning
strategy.

\textbf{Intrinsic images}.  It is natural to try to brighten shading while fixing
albedo. Wei et al.~\cite{Retinex-Net_2018}  decompose the input image into light-independent
reflectance and structure-aware smooth illumination components,
then adjust the illumination map for an enhanced
result.  Li et al.~\cite{LightenNet_2018} uses a four layer CNN,
LightenNet to estimate the illumination map. Zhang et
al.~\cite{KinD_2019} propose a more complex model, KinD consisting of
three subnetworks for layer decomposition, reflectance restoration,
and illumination adjustment. KinD++~\cite{KinD++_2021} is avoids the visual defects (e.g.,
non-uniform spots and oversmoothing) left in the outputs of KinD.
In contrast, we adjust images directly, with no estimate of albedo or illumination.
Intrinsic image methods typically ignore the enhanced noise of low light images,
leading to amplified noise in the output.  In contrast, our method uses
a model of low light noise in training to suppress output noise.

\textbf{Supervised learning}. Lore et al.~\cite{LLNet_2017}
apply an autoencoder (LLNet) to map a dark, noisy input image to a bright, denoised output
image. They show two strategies, simultaneous enhancement and
denoising, and sequential enhancement then denoising.   
Lv et al.~\cite{MBLLEN_2018} describe MBLLEN, a multi-branch framework
with extraction, enhancement, and fusion modules, each with a
convolutional neural network (CNN). Lu et
al.~\cite{TBEFN_2021} describe TBEFN, a two branch exposure-fusion
network without an end-to-end training scheme. Lim et
al.~\cite{DSLR_2021} describe a Laplacian pyramid based method (DSLR),
a stacked laplacian restorer to enhance images.  The primary difficulty with
such supervised learning methods is providing data at a large enough scale
to ensure generalization.  Methods described are limited by a shortage
of training images (see Tab.~\ref{table:quan_res}), dealt with by combining several real low light image datasets and using various augmentations. These approaches
tend to work well with the specific type of images used to train them,
but do not generalize well, likely due to limited scale of the training data.
In contrast, our approach generates arbitrary quantities of
\textbf{diverse} training data using simulation. 

\textbf{Unsupervised learning}.  Jiang et al.~\cite{EnlightenGAN_2021}  use an attention guided
UNet~\cite{unet_2015} as the generator with global-local discriminator
to handle spatially varying light conditions in input images.  Their
network accepts dark images, and brightens them because the adversary
sees natural light (bright) images. This method can introduce significant color deviations.  In
contrast, we use a pure supervised method, and have no color deviations.

\textbf{Semi-supervised learning}. Yang et al.~\cite{DRBN_2020}
propose a deep recursive band network (DRBN) which decomposes an image
using paired data then reconstructs without paired data using
perceptual quality driven adversarial learning.
Sharma et al.~\cite{nighttime_visibility_2021}  
boost the intensity in low light regions of night-time images while suppressing
noisy light effects (e.g., glow and glare). They decompose the input
image into low-frequency and high-frequency feature maps and process
them by two separate networks for light effects suppression and noise
removal,  respectively. Then they use a network to increase the
dynamic range of the processed low frequency feature maps and combine
with the high frequency feature maps to generate the enhanced output
image. Their model is trained with paired data, but uses non-reference
loss functions and so is partially unsupervised.   In
contrast, we use a pure supervised method (so have stable training),
and outperform these methods because we preserve color information by taking in images in LAB color space rather than RGB.

\textbf{Zero-shot learning}. 
ExCNet~\cite{ExCNet_2019} 
is an exposure correction network that directly estimates the best
fitting S-Curve for a given back-lit image.  Zhang et
al.~\cite{ExCNet_2019} use a loss function that maximizes the
visibility of all image blocks while keeping the relative diffference
between neighboring blocks; no paired data is required.
Zhu et al.~\cite{RRDNet_2020} propose a robust retinex decomposition network,
decomposing the input to three components: illumination, reflectance,
and noise, then recovering an image by adjusting the illumination map,
denoising the reflectance map, and combining.
In contrast, we use a pure supervised method, and have no artifacts and color shifts.

\section{Method}
\label{sec:method}
\subsection{Overview}
\begin{figure}
\centering
\includegraphics[width=4.8in]{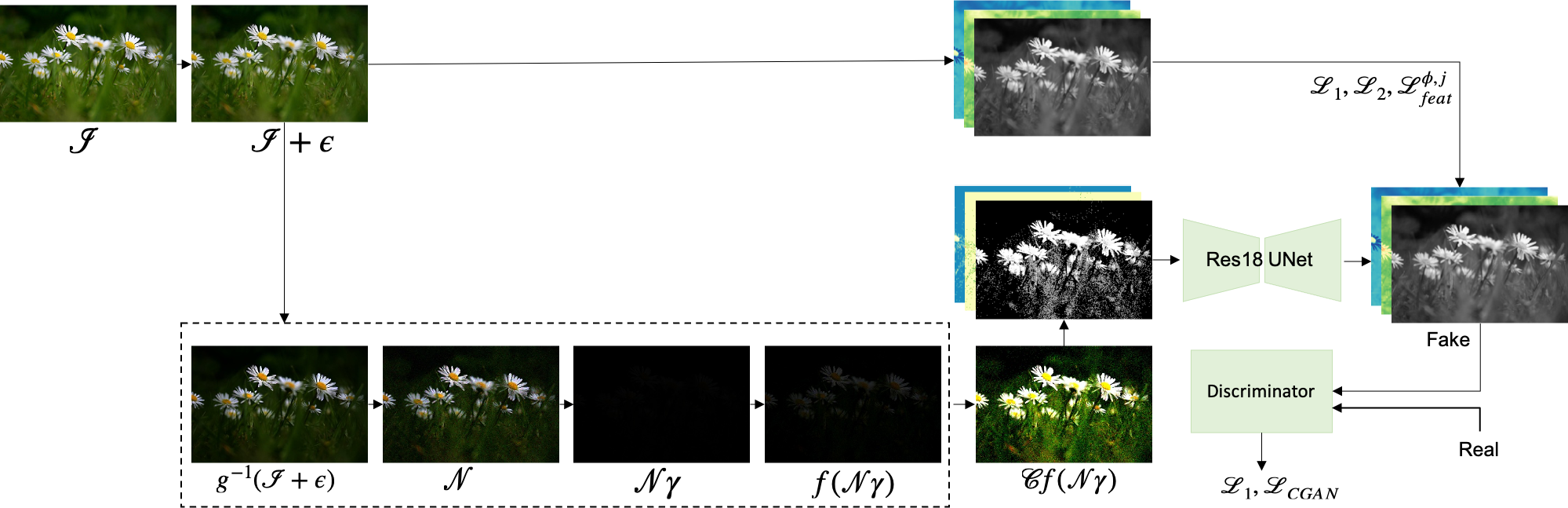}
\caption{The proposed method for low light image enhancement. We use a UNet with a pretrained ResNet18 backbone. There are two stages: (1) pretrain the network for few epochs with mean absolute error, mean squared error, and VGG perceptual loss; (2) train for additional epochs to ensure outputs result in realistic images imposed by an adversarial loss.} 
\label{fig:method}
\end{figure}

We propose an overall framework for generating synthetic data and end-to-end training of our convolutional network. We do not require RAW images. Unlike prior works, we operate on images in LAB color space and demonstrate great performance.
Fig.~\ref{fig:method} shows our proposed pipeline.

\subsection{Dataset}
We use a subset of 164k normal light images from the Places205~\cite{places205} dataset to simulate low light images. The dataset contains a variety of scenes and suitable for low light enhancement problem.

To produce a low light image, shown in Algorithm~\ref{alg:synth}, we take a normal image, and scale it by $1+\epsilon$ where $\epsilon$ is a random number fairly close to zero. We pass it to the low light procedure, shown in Algorithm~\ref{alg:lowlight}, where we pass the image through a known inverse camera response function (CRF) to produce a linearized image. Then, we add random shot and read noise to the linearized image and apply a small randomly chosen $\gamma$ weight to make it darker. We pass it through a known CRF to get the low light image. We compute a constant factor $k$ as the ratio of the mean of the bright image to the mean of the dark image, and scale the low light image by this constant. Now, we have a pair of images to train our supervised model.
Such procedure is quite useful as we generate training data on the fly. And based on our experimental findings, more data helps the supervised model to increase generalization capability.
\begin{algorithm}
	\caption{Low Light Model}\label{alg:lowlight}
	\begin{algorithmic}
	    \Procedure{lowLight}{image, $\gamma$}
    	    \State linearized $\gets$ \Call{$g^{-1}$}{image} \Comment{where $g^{-1}$ is known inverse CRF}
    	    \State add shot noise and read noise
    	    \State enhance darkness by $\gamma$
        \State \Return \Call{$f$}{noisy low light image} \Comment{where $f$ is known CRF}
        \EndProcedure
	\end{algorithmic}
\end{algorithm}
\begin{algorithm}
	\caption{Synthetic Image}\label{alg:synth}
	\begin{algorithmic}
	    \Procedure{synthesize}{image}
    	    \State H $\gets$ \Call{}{1+$\epsilon$}image \Comment{where $\epsilon \in$ random.uniform(-0.1,0.1)}
    	    
    	    \State $low_{H}$ $\gets$ \Call{lowlight}{H,$\gamma$} \Comment{where $\gamma$ $\in$ random.uniform(0.01,0.09)} 
    	    
    	    \State k $\gets$ $\frac{\Call{mean}{H}}{\Call{mean}{low_{H}}}$
    	    
    	    \State L $\gets$ \Call{}{k}$low_H$

            \State \Return \Call{}{H, L} \Comment{model inputs}
        \EndProcedure
	\end{algorithmic}
\end{algorithm}

\subsection{Model}
\textbf{Network}. After preliminary exploration, we focus on an UNet~\cite{unet_2015} with ResNet18~\cite{resnet_2015} backbone generator and a patch discriminator~\cite{im2im_2018}.
Our model takes an input of size 256x256x3 in LAB color space, and outputs the predicted bright image. We use several losses for supervision to ensure the network learns the mapping from a dark image to bright image.

\textbf{Supervision.} 
We pretrain the UNet model on the training dataset with L1 loss, L2 loss, and VGG perceptual loss for a few epochs. 
Then we train the whole model on the training dataset with combined adversarial loss and L1 loss. 

 Thus, we consider VGG perceptual loss~\cite{vgg_perceptial_loss_2016}
Similar to the existing works, we use L1 and L2 losses; however, they alone are not enough, resulting in outputs with artifacts or major color shifts.  They focus on low level information in the image. To consider higher level features, we consider using VGG19 perceptual loss ~\cite{vgg_perceptial_loss_2016} as follows:
\begin{equation}
    \mathcal{L}_{feat}^{\phi,j}(\hat{y},y) = \frac{1}{c_{j}h_{j}w_{j}} ||\phi_{j}(\hat{y}) - \phi_{j}(y) ||_{1}^{1}
    \label{eq:vgg}
\end{equation}
where $\phi_{j}$ denotes the VGG features at j-th layer, $y$ and $\hat{y}$ are the ground truth and predicted output, respectively; $c,h,w$ each represent channel, height, and width of the image.

To further improve generalization, we include an adversary to help solve the problem in an unsupervised manner. We use a conditional adversarial network~\cite{im2im_2018} where the discriminator sees the fake outputs as well as some real inputs to learn which is which.
More precisely, let us consider $x$ as the low light image and $y$ as the normal light image. $G$ is the generator model and $D$ is the discriminator model. The loss is: 
\begin{equation}
    \mathcal{L}_{cGAN} (G,D) = \mathbb{E}_{y}[log D(y)]+ \mathbb{E}_{x}[log(1 - D(G(x)))]
    \label{eq:adv}
\end{equation}

Eq.~\ref{eq:adv} helps generate real looking results, but to help the model and introduce some supervision, we combine it with L1 loss, mean absolute error of the prediction $G(x)$ and the target $y$
\begin{equation}
    \label{eqadvl1}
    \mathcal{L}_{1}(G) = E_{x,y}[||y - G(x) ||_{1}]
\end{equation}

Using only L1 loss makes the model to be conservative and mostly show saturated results if in doubt which color is correct when mapping from the dark input image to the bright output image. This is because it takes an average and tries to reduce L1 loss as much as possible. Thus, the combined loss function is: 
\begin{equation}
    G* = arg \underset{G}{\mathrm{min}}\underset{D}{\mathrm{max }} \mathcal{L}_{cGAN}(G,D)+ \lambda \mathcal{L}_{1}(G) 
\end{equation}

where $\lambda$ balances the contribution of the two losses to the final loss. 

\subsection{Implementation Details}
Our implementation is done in PyTorch. The proposed model is relatively quick to converge. The model we use in Sec.~\ref{sec:experiment} is trained for a total of 8 epochs. The generator is pretrained for 4 epochs. And the generator and discriminator are trained for an additional 4 epochs. There are a total of 164k images, with 16k for training and 4k for testing. We use a batch size of 32 of 256x256x3 inputs. The images are in LAB color space and scaled to -1 to 1.The training is done using ADAM optimizer with learning rate $\alpha=2e-4$, $\beta_{1} = 0.5$ and $\beta_{2} = 0.999$. The balancing term $\lambda=100$.

\section{Experiments}
\label{sec:experiment}
\subsection{Setup}
We perform the following experiments. 1) We evaluate the proposed method on our Places205~\cite{places205} synthetic test-set images. 2) We compare quantitatively and qualitatively with representative works in non RAW image format on the task of low light image enhancement on two paired standard datasets, LOL~\cite{LOL_2018} test-set and MIT-Adobe-FiveK~\cite{fivek} test-set. 3) We compare on the challenging LLIV-Phone~\cite{survey_2021} test-set. Please note it only contains low light images with no reference ground truth bright images.

\textbf{Evaluation dataset}. Following the recent survey~\cite{survey_2021} to test the generalization capability of our method and baseline methods, we adopt the commonly used LOL~\cite{LOL_2018} test-set, MIT-Adobe-FiveK~\cite{fivek} test-set, and LLIV-Phone~\cite{survey_2021} test-set. The entire LOL dataset contains 500 images at 400x600x3 resolutions. The test-set contains only 15 images. The entire   MIT-Adobe-FiveK consists of 5000 images in RAW format. The test-set contains only 500 images. We follow the survey~\cite{survey_2021} procedure to process the data. We use Adobe Lightroom to resize the images to have a long edge of 512 pixels and export them in PNG format. 
The LLIV-Phone consists of 45,184 images taken by 18 different mobile phones under different illumination conditions. Again, we follow the survey~\cite{survey_2021} and select 600 random images at half of the original resolution, denoted as LLIV-Phone-imgT.

\textbf{Evaluation metrics.}
We use three reference based widely used image quality assessment metrics. Peak-signal-noise-ratio (PSNR) where the greater the value is the better the model prediction is. Structural similarity index measure (SSIM)~\cite{ssim} is a perception based image quality assessment metric that considers image degredation as perceived change in structural information. The closer the value is to 1, the better the model prediction is. LPIPS~\cite{lpips} , a deep learning based image quality assessment metric that measures the perceptual similarity by deep visual representations between the predicted output and its corresponding ground truth. The smaller value suggests closer perceptual similarity between the prediction and the target.  Following ~\cite{survey_2021}, we use the AlexNet-based model to compute perceptual similarity.
We use two non-reference based image quality assessment metrics, NIQE~\cite{niqe} and SPAQ~\cite{spaq} to quantitatively compare our method with the other baselines on the LLIV-Phone-ImgT dataset. Smaller values for NIQE mean better visual quality. Larger values for SPAQ mean better perceptual quality.  

\textbf{Baseline methods}. We compare the proposed method with four categories of methods: 1) supervised learning including LLNet~\cite{LLNet_2017}, LightenNet~\cite{LightenNet_2018}, Retinex-Net~\cite{Retinex-Net_2018}, MBLLEN~\cite{MBLLEN_2018}, KinD~\cite{KinD_2019}, KinD++~\cite{KinD++_2021}, TBEFN~\cite{TBEFN_2021}, and DSLR~\cite{DSLR_2021}, 2) unsupervised learning, i.e., EnlightenGAN~\cite{EnlightenGAN_2021}, 3) semi supervised learning, i.e., DRBN~\cite{DRBN_2020}, and 4) zero shot learning including ExCNet~\cite{ExCNet_2019}, Zero-DCE~\cite{Zero-DCE_2019}, and RRDNet~\cite{RRDNet_2020}. A comprehensive review of these recent methods and their results are provided in the recent survey~\cite{survey_2021}. They use the publicly available code for each method to produce their results for fair comparisons. We rely on their benchmarking results, and use them for our comparisons.
\subsection{Results}
\begin{table}[ht!]
\begin{center}
\caption{Quantitative comparisons in terms of reference based metrics( PSNR (in dB), SSIM, LPIPS) and non-reference based metrics (NIQE and SPAQ) on standard datasets.  We did not use any of these datasets in training, in contrast to common practice. There is a check mark for methods that {\bf use} LOL or MIT in training. The best result is in black bold, second best in blue, third best in green.  Note: Supervised methods are dominant; our method has an order of magnitude more training data than any other; our method is best in all metrics for LOL, and very strong for MIT-Adobe-FiveK; our method is weaker in NIQE and SPAQ metrics, but we place less weight on non-reference metrics (which do not have access to ground truth when assessing a result). S stands for supervised, U for unsupervised, SS for semi-supervised, and ZS for zero-shot.}
\label{table:quan_res}
\resizebox{\textwidth}{!}{\begin{tabular}{| c | c | r | c | c | r | r | r | r | r | r | r | r |}
\cline{3-13}
\multicolumn{2}{c|}{} & \multicolumn{3}{c|}{Train Data} & \multicolumn{8}{c|}{Test Data}\\
\cline{6-13}
\multicolumn{2}{c|}{} & \multicolumn{3}{c|}{} & \multicolumn{3}{c|}{LOL} & \multicolumn{3}{c|}{MIT-Adobe-FiveK} & \multicolumn{2}{c|}{LLIV-Phone-ImgT}\\
\hline
Type & Methods & \# Images & LOL & MIT & {PSNR}$\uparrow$ & {SSIM}$\uparrow$ & {LPIPS}$\downarrow$  & {PSNR}$\uparrow$ & {SSIM}$\uparrow$ & {LPIPS}$\downarrow$ & {NIQE}$\downarrow$ & {SPAQ}$\uparrow$\\
\hline
S & {LLNet~\cite{LLNet_2017}} & 485 & \checkmark & & \textcolor{blue}{17.959} & 0.713 & 0.360 & 12.177 & 0.645 & 0.292 & 5.86 & 40.56\\
& {LightenNet~\cite{LightenNet_2018}} & 600 & & & 10.301 & 0.402 & 0.394 & 13.579 & 0.744 & \textcolor{green}{0.166} & 5.34 & 45.74\\
& {Retinex-Net~\cite{Retinex-Net_2018}} & 1485 & \checkmark & & 16.774 & 0.462 & 0.474 & 12.310 & 0.671 & 0.239 & 5.01 & \textbf{50.95}\\
& {MBLLEN~\cite{MBLLEN_2018}} & 16925 & & & \textcolor{green}{17.902} & 0.715 & 0.247 & \textcolor{blue}{19.781} & \textbf{0.825} & \textbf{0.108} & 5.08 & 42.50\\
& {KinD~\cite{KinD_2019}} & 485 & \checkmark & & 17.648 & \textcolor{green}{0.779} & \textcolor{blue}{0.175} & 14.535 & 0.741 & 0.177 & 4.97 & 44.79\\
& {KinD++~\cite{KinD++_2021}} & 725 & \checkmark & & 17.752 & 0.760 & \textcolor{green}{0.198} & 9.732 & 0.568 & 0.336 & \textbf{4.73} & \textcolor{blue}{46.89}\\
& {TBEFN~\cite{TBEFN_2021}} & 1074 & \checkmark & & 17.351 & \textcolor{blue}{0.786} & 0.210 & 12.769 & 0.704 & 0.178 & 4.81 & 44.14\\
& {DSLR~\cite{DSLR_2021}} & 4500 & & \checkmark & 15.050 & 0.597 & 0.337 & \textcolor{green}{16.632} & \textcolor{blue}{0.782} & 0.167 & \textcolor{blue}{4.77} & 41.08\\
& Ours & 160000 & & & \textbf{21.485} & \textbf{0.820} & \textbf{0.057} & \textbf{20.344} & \textcolor{green}{0.754} & \textcolor{blue}{0.109} & 5.17 & 45.14 \\
\hline
U & {EnlightenGAN~\cite{EnlightenGAN_2021}} & 1930 &  & & 17.483 & 0.677 & 0.322 & 13.260 & 0.745 & 0.170 & \textcolor{green}{4.79} & 45.48\\
\hline
SS & {DRBN~\cite{DRBN_2020}} & 485 & \checkmark & & 15.125 & 0.472 & 0.316 & 13.355 & 0.378 & 0.281 & 5.80 & 42.74\\
\hline
ZS & {ExCNet~\cite{ExCNet_2019}} & 1000 & & & 15.783 & 0.515 & 0.373 & 13.978 & 0.710 & 0.187 & 5.55 & 46.74\\
& {Zero-DCE~\cite{Zero-DCE_2019}} & 2422 & & & 14.861 & 0.589 & 0.335 & 13.199 & 0.709 & 0.203 & 5.82 & \textcolor{green}{46.85}\\
& {RRDNet~\cite{RRDNet_2020}} & - &  & & 11.392 & 0.468 & 0.361 & 10.135 & 0.620 & 0.303 & 5.97 & 45.31\\
\hline
\end{tabular}}
\end{center}
\end{table}

\textbf{Quantitative results}. 
Tab.~\ref{table:quan_res} shows that our proposed method outperforms other baselines by a great margin in LOL test-set. It achieves the highest PSNR on the MIT-Adobe-FiveK dataset, but slightly lower than MBLLEN~\cite{MBLLEN_2018} in terms of SSIM and LPIPS.
In Tab.~\ref{table:quan_res} we report the NIQE and SPAQ scores on the LLIV-Phone-ImgT dataset. As shown, our method outperforms some of the methods including LLNet~\cite{LLNet_2017}, LightenNet~\cite{LightenNet_2018}, DRBN~\cite{DRBN_2020}, ExCNet~\cite{ExCNet_2019}, Zero-DCE~\cite{Zero-DCE_2019}, and RRDNet~\cite{RRDNet_2020}; however, it achieves slightly lower than comparing methods RetinexNet~\cite{Retinex-Net_2018}, MBLLEN~\cite{MBLLEN_2018}, KindD~\cite{KinD_2019}, KinD++~\cite{KinD++_2021}, TBEFN~\cite{TBEFN_2021}, DLSR~\cite{DSLR_2021}, and EnlightenGAN~\cite{EnlightenGAN_2021}.

For non-reference image quality metrics where we don't have the bright image available, we select the constant $k$ as follows.
We take our training dataset and compute its mean intensity. We select the constant such that if we multiply the input by that constant, we get the mean intensity. Intuitively, what is the brightness of the output image if we don't know the output image (i.e. reference image)? It is the mean brightness of the images. Thus, our best estimate is the mean intensity of all the bright images the model has seen.

\begin{figure}
\scriptsize
\stackunder[1pt]{\includegraphics[width=0.24\textwidth]{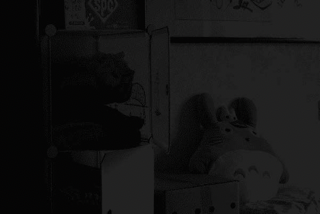}}{Input}
\stackunder[1pt]{\includegraphics[width=0.24\textwidth]{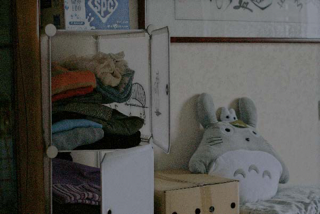}}{Zero-DCE~\cite{Zero-DCE_2019}}
\stackunder[1pt]{\includegraphics[width=0.24\textwidth]{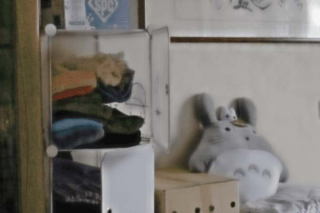}}{LLNet~\cite{LLNet_2017} }
\stackunder[1pt]{\includegraphics[width=0.24\textwidth]{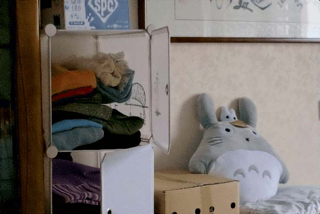}}{MBLLEN~\cite{MBLLEN_2018}}
\stackunder[1pt]{\includegraphics[width=0.24\textwidth]{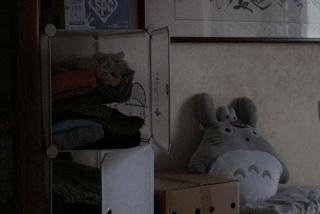}}{LightenNet~\cite{LightenNet_2018}}
\stackunder[1pt]{\includegraphics[width=0.24\textwidth]{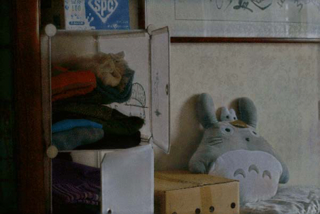}}{DSLR~\cite{DSLR_2021}}
\stackunder[1pt]{\includegraphics[width=0.24\textwidth]{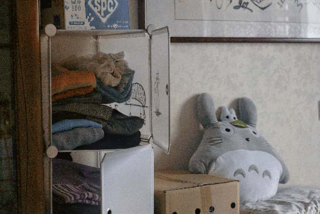}}{EnlightenGAN~\cite{EnlightenGAN_2021}}
\stackunder[1pt]{\includegraphics[width=0.24\textwidth]{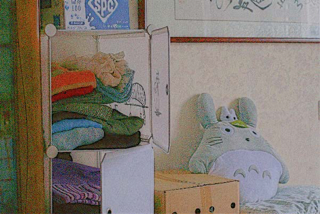}}{Retinex-Net~\cite{Retinex-Net_2018}}
\stackunder[1pt]{\includegraphics[width=0.24\textwidth]{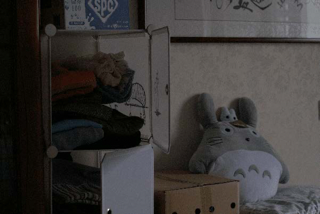}}{RRDNet~\cite{RRDNet_2020}}
\stackunder[1pt]{\includegraphics[width=0.24\textwidth]{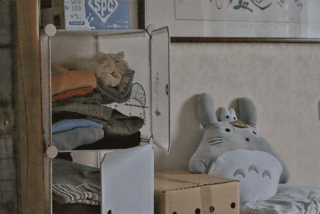}}{TBEFN~\cite{TBEFN_2021}}
\stackunder[1pt]{\includegraphics[width=0.24\textwidth]{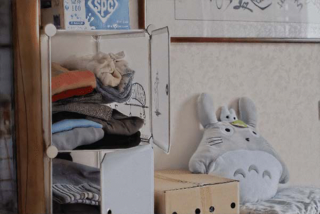}}{KinD~\cite{KinD_2019}}
\stackunder[1pt]{\includegraphics[width=0.24\textwidth]{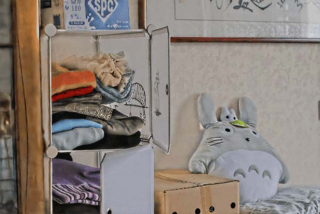}}{KinD++~\cite{KinD++_2021}}
\stackunder[1pt]{\includegraphics[width=0.24\textwidth]{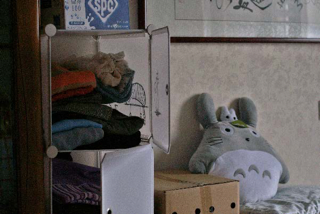}}{ ExCNet~\cite{ExCNet_2019}}
\stackunder[1pt]{\includegraphics[width=0.24\textwidth]{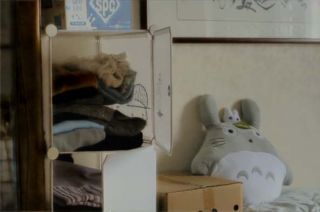}}{DRBN~\cite{DRBN_2020}}
\stackunder[1pt]{\includegraphics[width=0.24\textwidth]{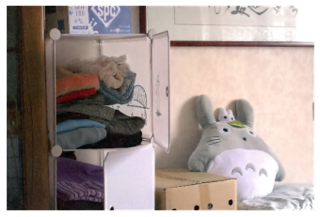}}{Ours}
\stackunder[1pt]{\includegraphics[width=0.24\textwidth]{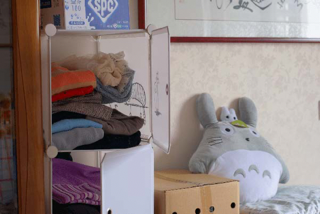}}{Ground Truth}
\stackunder[1pt]{\includegraphics[width=0.24\textwidth]{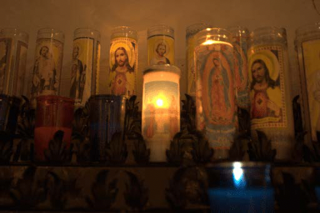}}{Input}
\stackunder[1pt]{\includegraphics[width=0.24\textwidth]{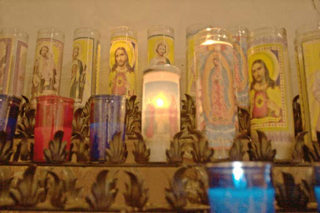}}{Zero-DCE~\cite{Zero-DCE_2019}}
\stackunder[1pt]{\includegraphics[width=0.24\textwidth]{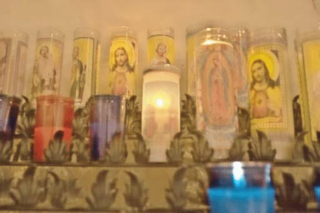}}{LLNet~\cite{LLNet_2017} }
\stackunder[1pt]{\includegraphics[width=0.24\textwidth]{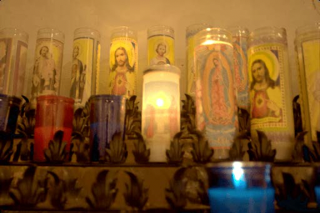}}{MBLLEN~\cite{MBLLEN_2018}}
\stackunder[1pt]{\includegraphics[width=0.24\textwidth]{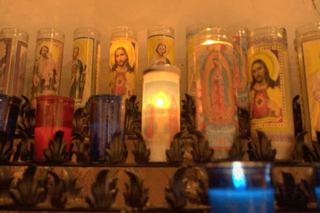}}{LightenNet~\cite{LightenNet_2018}}
\stackunder[1pt]{\includegraphics[width=0.24\textwidth]{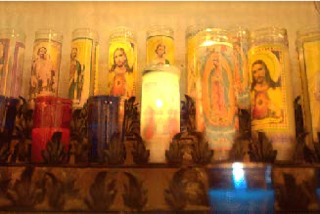}}{DSLR~\cite{DSLR_2021}}
\stackunder[1pt]{\includegraphics[width=0.24\textwidth]{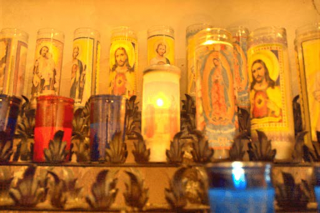}}{EnlightenGAN~\cite{EnlightenGAN_2021}}
\stackunder[1pt]{\includegraphics[width=0.24\textwidth]{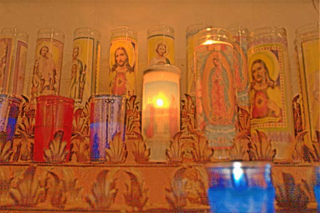}}{Retinex-Net~\cite{Retinex-Net_2018}}
\stackunder[1pt]{\includegraphics[width=0.24\textwidth]{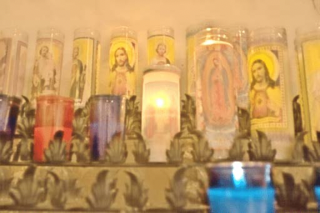}}{RRDNet~\cite{RRDNet_2020}}
\stackunder[1pt]{\includegraphics[width=0.24\textwidth]{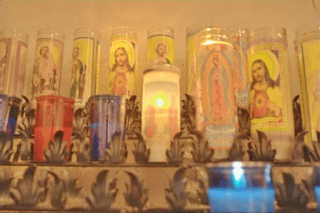}}{TBEFN~\cite{TBEFN_2021}}
\stackunder[1pt]{\includegraphics[width=0.24\textwidth]{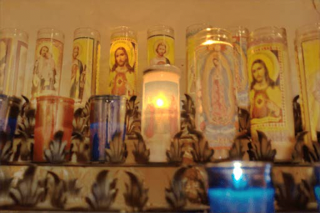}}{KinD~\cite{KinD_2019}}
\stackunder[1pt]{\includegraphics[width=0.24\textwidth]{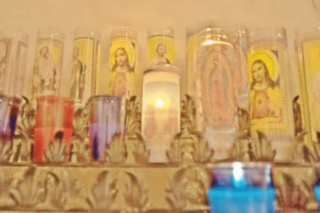}}{KinD++~\cite{KinD++_2021}}
\stackunder[1pt]{\includegraphics[width=0.24\textwidth]{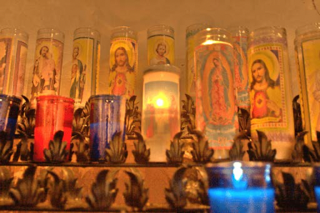}}{ ExCNet~\cite{ExCNet_2019}}
\stackunder[1pt]{\includegraphics[width=0.24\textwidth]{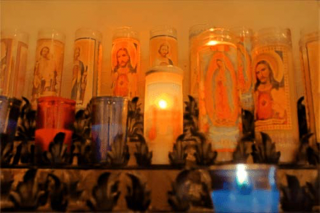}}{DRBN~\cite{DRBN_2020}}
\stackunder[1pt]{\includegraphics[width=0.24\textwidth]{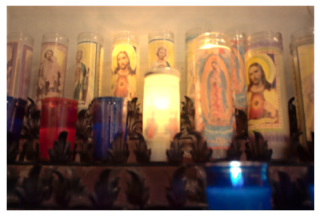}}{Ours}
\stackunder[1pt]{\includegraphics[width=0.24\textwidth]{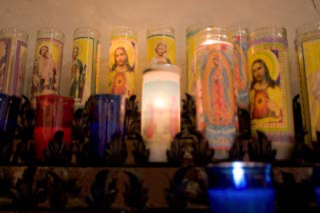}}{Ground Truth}
\caption{Results on images from LOL (top) and MIT-Adobe-FiveK (bottom)}
\label{fig:qual1-2}
\end{figure}

\begin{figure}
\scriptsize
\stackunder[1pt]{\includegraphics[width=0.24\textwidth]{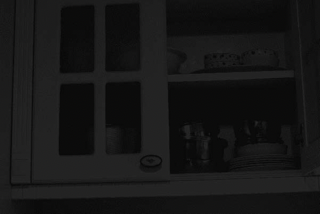}}{Input}
\stackunder[1pt]{\includegraphics[width=0.24\textwidth]{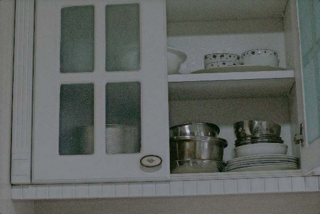}}{Zero-DCE~\cite{Zero-DCE_2019}}
\stackunder[1pt]{\includegraphics[width=0.24\textwidth]{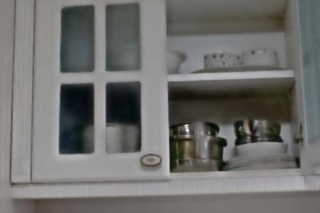}}{LLNet~\cite{LLNet_2017} }
\stackunder[1pt]{\includegraphics[width=0.24\textwidth]{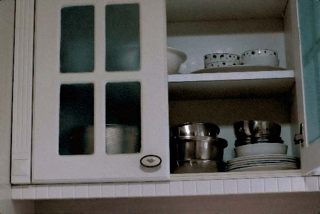}}{MBLLEN~\cite{MBLLEN_2018}}
\stackunder[1pt]{\includegraphics[width=0.24\textwidth]{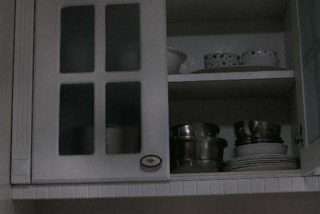}}{LightenNet~\cite{LightenNet_2018}}
\stackunder[1pt]{\includegraphics[width=0.24\textwidth]{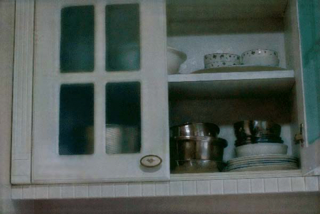}}{DSLR~\cite{DSLR_2021}}
\stackunder[1pt]{\includegraphics[width=0.24\textwidth]{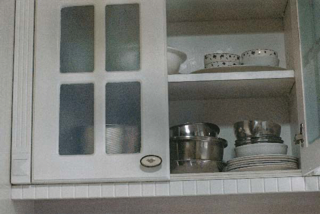}}{EnlightenGAN~\cite{EnlightenGAN_2021}}
\stackunder[1pt]{\includegraphics[width=0.24\textwidth]{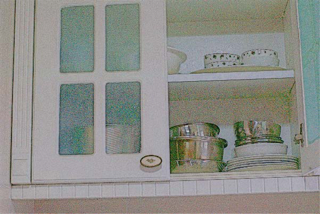}}{Retinex-Net~\cite{Retinex-Net_2018}}
\stackunder[1pt]{\includegraphics[width=0.24\textwidth]{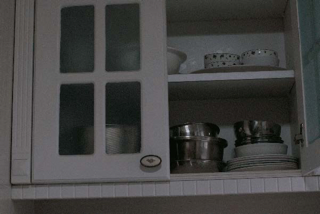}}{RRDNet~\cite{RRDNet_2020}}
\stackunder[1pt]{\includegraphics[width=0.24\textwidth]{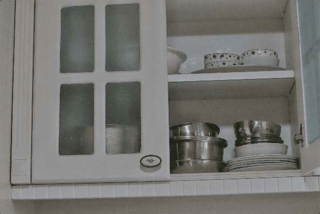}}{TBEFN~\cite{TBEFN_2021}}
\stackunder[1pt]{\includegraphics[width=0.24\textwidth]{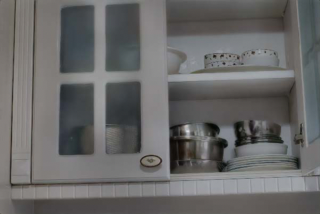}}{KinD~\cite{KinD_2019}}
\stackunder[1pt]{\includegraphics[width=0.24\textwidth]{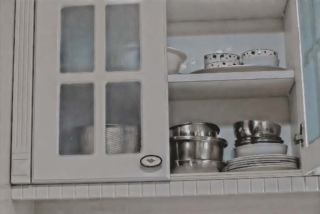}}{KinD++~\cite{KinD++_2021}}
\stackunder[1pt]{\includegraphics[width=0.24\textwidth]{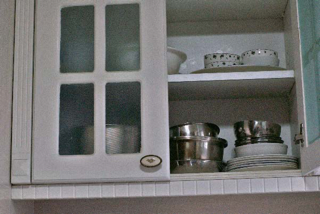}}{ ExCNet~\cite{ExCNet_2019}}
\stackunder[1pt]{\includegraphics[width=0.24\textwidth]{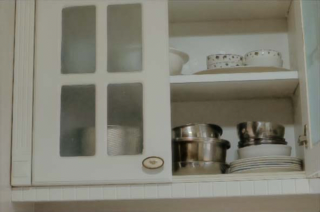}}{DRBN~\cite{DRBN_2020}}
\stackunder[1pt]{\includegraphics[width=0.24\textwidth]{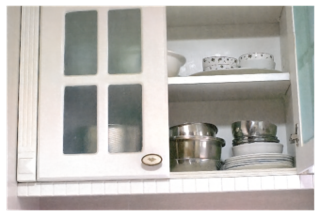}}{Ours}
\stackunder[1pt]{\includegraphics[width=0.24\textwidth]{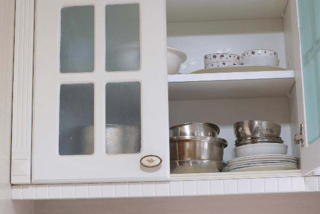}}{Ground Truth}
\stackunder[1pt]{\includegraphics[width=0.24\textwidth]{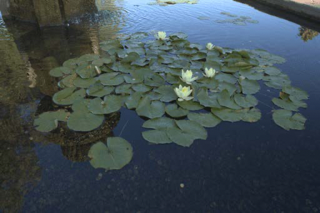}}{Input}
\stackunder[1pt]{\includegraphics[width=0.24\textwidth]{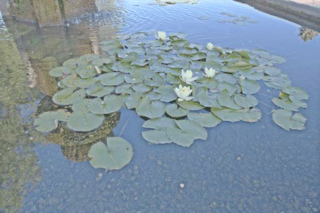}}{Zero-DCE~\cite{Zero-DCE_2019}}
\stackunder[1pt]{\includegraphics[width=0.24\textwidth]{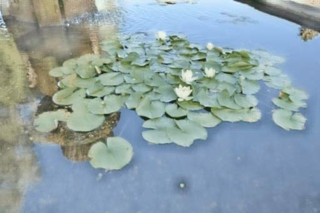}}{LLNet~\cite{LLNet_2017} }
\stackunder[1pt]{\includegraphics[width=0.24\textwidth]{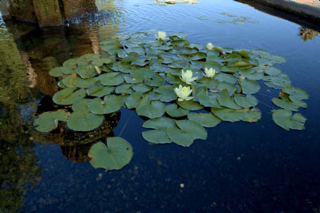}}{MBLLEN~\cite{MBLLEN_2018}}
\stackunder[1pt]{\includegraphics[width=0.24\textwidth]{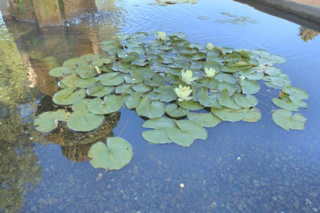}}{LightenNet~\cite{LightenNet_2018}}
\stackunder[1pt]{\includegraphics[width=0.24\textwidth]{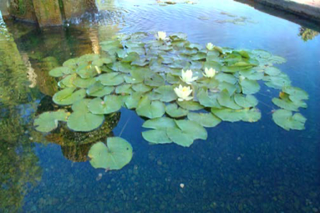}}{DSLR~\cite{DSLR_2021}}
\stackunder[1pt]{\includegraphics[width=0.24\textwidth]{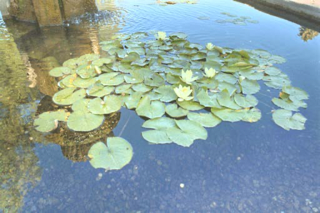}}{EnlightenGAN~\cite{EnlightenGAN_2021}}
\stackunder[1pt]{\includegraphics[width=0.24\textwidth]{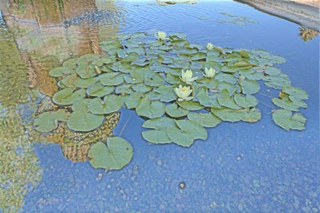}}{Retinex-Net~\cite{Retinex-Net_2018}}
\stackunder[1pt]{\includegraphics[width=0.24\textwidth]{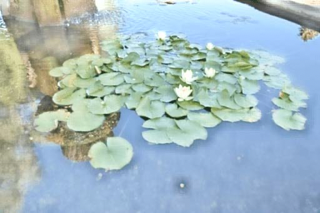}}{RRDNet~\cite{RRDNet_2020}}
\stackunder[1pt]{\includegraphics[width=0.24\textwidth]{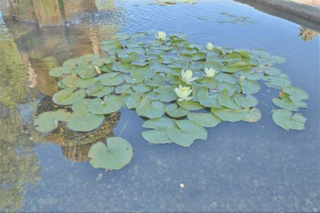}}{TBEFN~\cite{TBEFN_2021}}
\stackunder[1pt]{\includegraphics[width=0.24\textwidth]{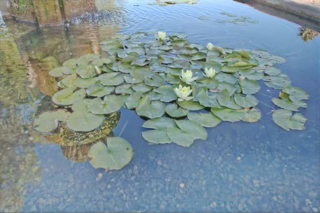}}{KinD~\cite{KinD_2019}}
\stackunder[1pt]{\includegraphics[width=0.24\textwidth]{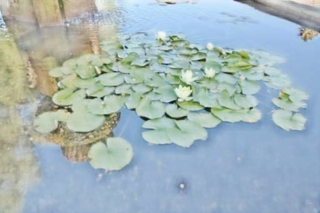}}{KinD++~\cite{KinD++_2021}}
\stackunder[1pt]{\includegraphics[width=0.24\textwidth]{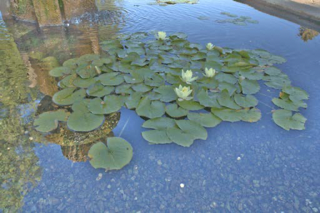}}{ ExCNet~\cite{ExCNet_2019}}
\stackunder[1pt]{\includegraphics[width=0.24\textwidth]{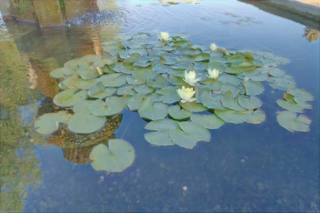}}{DRBN~\cite{DRBN_2020}}
\stackunder[1pt]{\includegraphics[width=0.24\textwidth]{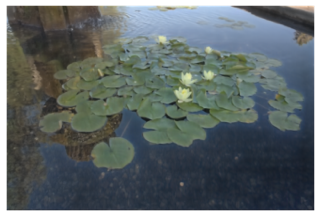}}{Ours}
\stackunder[1pt]{\includegraphics[width=0.24\textwidth]{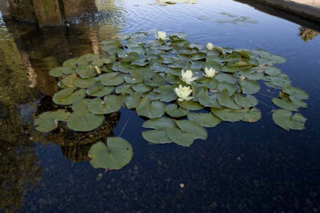}}{Ground Truth}
\caption{Results on images from LOL (top) and MIT-Adobe-FiveK (bottom)}
\label{fig:qual3-4}
\end{figure}

\textbf{Qualitative results}. Figs.~\ref{fig:qual1-2} -~\ref{fig:qual3-4} show some qualitative results on the LOL and MIT-Adobe-FiveK test-sets.
For example, Fig.~\ref{fig:qual1-2} (top) shows our result is close to the ground truth image. The overall brightness and colors are similar, but the wood shown in the left is a bit darker color in ours versus the ground truth. Similarly, none of the methods correctly recover the color of the violet piece of clothing in the left. As another example, Fig.~\ref{fig:qual3-4} (top) shows methods do not well improve the brightness of the low light input image. Our predicted output is similar to the ground truth image. 

Fig.~\ref{fig:qual1-2} (bottom) shows methods improve brightness of the low light input image. However, some methods produce either overexposed or underexposed results. Our predicted output is closer to the ground truth specifically when compared to the other outputs. For instance, the blue glass in the bottom-right corner or the red and blue glasses in the top-left corner are most similar to the ground truth than results of the other methods. Similarly, Fig.~\ref{fig:qual3-4} (bottom) shows our method produces the closest bright image to the ground truth.

\begin{figure}
\scriptsize
\stackunder[1pt]{\includegraphics[width=0.24\textwidth]{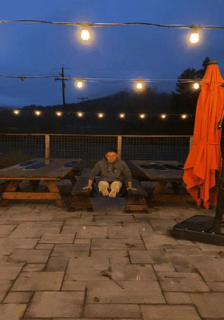}}{Input}
\stackunder[1pt]{\includegraphics[width=0.24\textwidth]{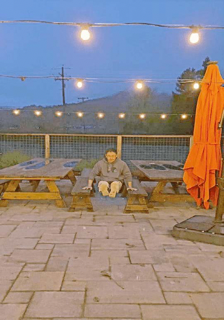}}{Zero-DCE~\cite{Zero-DCE_2019}}
\stackunder[1pt]{\includegraphics[width=0.24\textwidth]{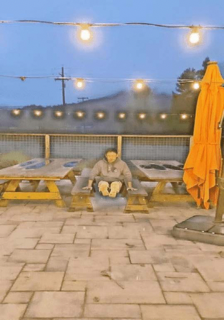}}{LLNet~\cite{LLNet_2017} }
\stackunder[1pt]{\includegraphics[width=0.24\textwidth]{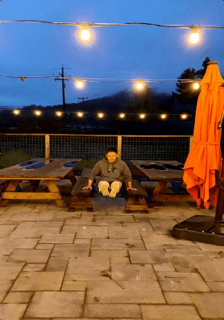}}{MBLLEN~\cite{MBLLEN_2018}}
\stackunder[1pt]{\includegraphics[width=0.24\textwidth]{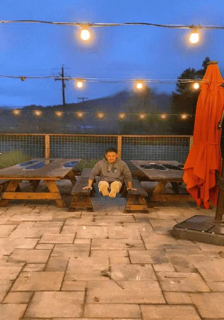}}{LightenNet~\cite{LightenNet_2018}}
\stackunder[1pt]{\includegraphics[width=0.24\textwidth]{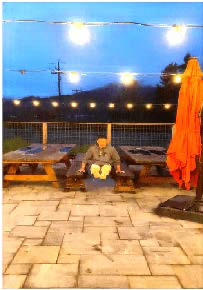}}{DSLR~\cite{DSLR_2021}}
\stackunder[1pt]{\includegraphics[width=0.24\textwidth]{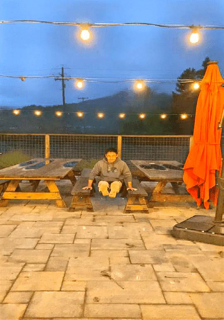}}{EnlightenGAN~\cite{EnlightenGAN_2021}}
\stackunder[1pt]{\includegraphics[width=0.24\textwidth]{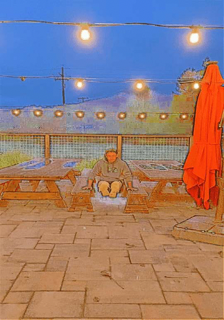}}{Retinex-Net~\cite{Retinex-Net_2018}}
\stackunder[1pt]{\includegraphics[width=0.24\textwidth]{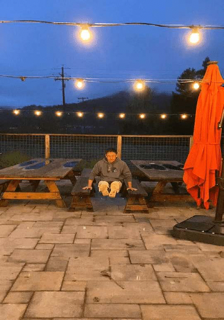}}{RRDNet~\cite{RRDNet_2020}}
\stackunder[1pt]{\includegraphics[width=0.24\textwidth]{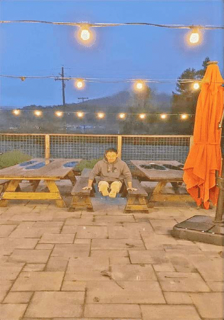}}{TBEFN~\cite{TBEFN_2021}}
\stackunder[1pt]{\includegraphics[width=0.24\textwidth]{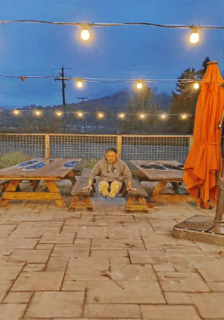}}{KinD~\cite{KinD_2019}}
\stackunder[1pt]{\includegraphics[width=0.24\textwidth]{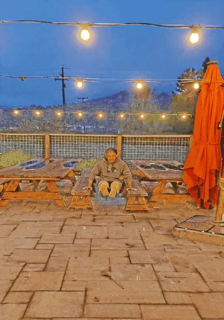}}{KinD++~\cite{KinD++_2021}}
\stackunder[1pt]{\includegraphics[width=0.24\textwidth]{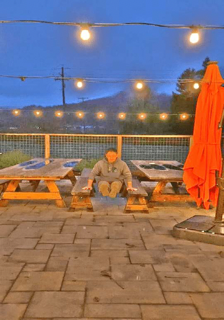}}{ ExCNet~\cite{ExCNet_2019}}
\stackunder[1pt]{\includegraphics[width=0.24\textwidth]{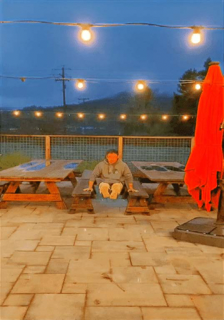}}{DRBN~\cite{DRBN_2020}}
\stackunder[1pt]{\includegraphics[width=0.24\textwidth]{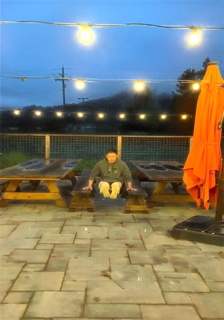}}{Ours}
\caption{An example result on LLIV-Phone-ImgT test-set - note there is no ground truth.}
\label{fig:qual-phone}
\end{figure}

Fig.~\ref{fig:qual-phone} shows results of different methods on an input sample from LLIV-Phone-ImgT. As can be seen, all methods exaggerate contrast or overexpose except MBLLEN. But it darkens the back of the scene. RRDNet and our output produce an appealing bright image from the input dark image.

\subsection{Ablations and Limitations}
\begin{figure*}[!ht]
	\centering
	\begin{subfigure}{0.16\textwidth}
		\includegraphics[width=\textwidth]{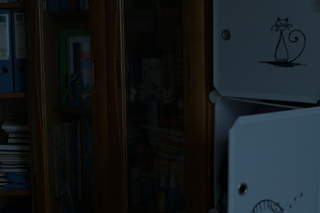}
	\end{subfigure}
	\begin{subfigure}{0.16\textwidth}
		\includegraphics[width=\textwidth]{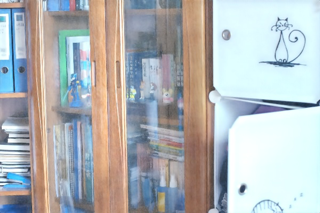}
	\end{subfigure}
	\begin{subfigure}{0.16\textwidth}
		\includegraphics[width=\textwidth]{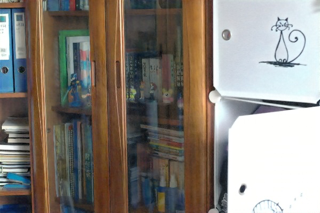}
	\end{subfigure}
	\begin{subfigure}{0.16\textwidth}
		\includegraphics[width=\textwidth]{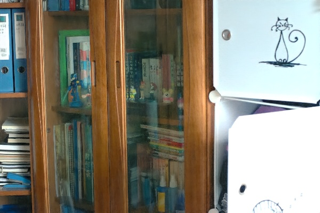}
	\end{subfigure}
	\begin{subfigure}{0.16\textwidth}
		\includegraphics[width=\textwidth]{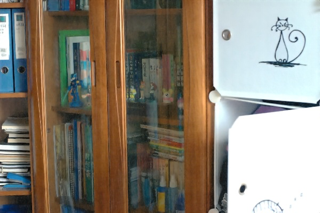}
	\end{subfigure}
	\begin{subfigure}{0.16\textwidth}
		\includegraphics[width=\textwidth]{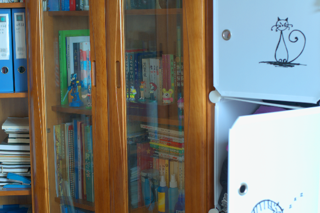}
	\end{subfigure}
	
	\begin{subfigure}{0.16\textwidth}
		\includegraphics[width=\textwidth]{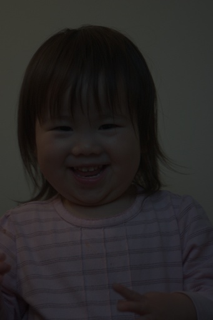}
		\caption{ \label{fig:ab_mit_in}}
	\end{subfigure}
	\begin{subfigure}{0.16\textwidth}
		\includegraphics[width=\textwidth]{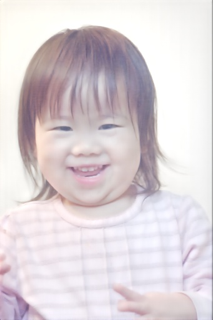}
		\caption{ \label{fig:ab_mit_out6}}
	\end{subfigure}
	\begin{subfigure}{0.16\textwidth}
		\includegraphics[width=\textwidth]{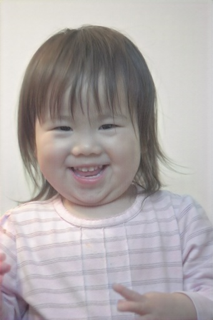}
		\caption{\label{fig:ab_mit_out4}}
	\end{subfigure}
	\begin{subfigure}{0.16\textwidth}
		\includegraphics[width=\textwidth]{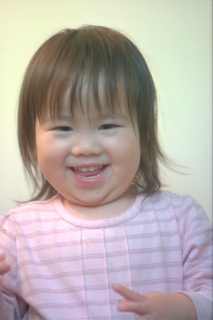}
		\caption{ \label{fig:ab_mit_out5}}
	\end{subfigure}
	\begin{subfigure}{0.16\textwidth}
		\includegraphics[width=\textwidth]{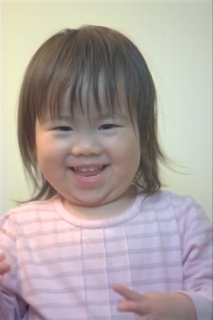}
		\caption{ \label{fig:ab_mit_outp}}
	\end{subfigure}
	\begin{subfigure}{0.16\textwidth}
		\includegraphics[width=\textwidth]{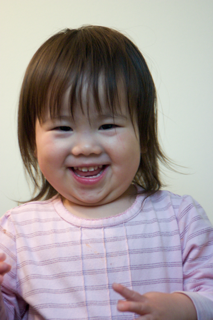}
		\caption{ \label{fig:ab_mit_gt}}
	\end{subfigure}
	
	\caption{\label{fig:ablation} Results of different ablations on an input sample from LOL~\cite{LOL_2018} test-set(top) and MIT-Adobe-FiveK~\cite{fivek} (bottom). (a) Input. (b) without multiplying input by constant $k$. (c) without converting input to LAB color space. (d) without adversarial loss. (e) output of the proposed solution. (f) ground truth. }
\end{figure*}

To analyze our method, we perform a series of ablation studies to determine how different design choices affect performance. We verify the results on our testing dataset, LOL~\cite{LOL_2018} and MIT-Adobe-FiveK~\cite{fivek} that were not seen by our model at training time. Tab.~\ref{table:ablations} summarizes the quantitative results and Fig.~\ref{fig:ablation} shows relevant qualitative results. Please note we look at each design choice one at a time. For example, we remove noise augmentation and keep everything else.

{\bf w/o $1+\epsilon$}. 
Generating low light images by ablating $1+\epsilon$ augmentation means we pass in the original normal light images to the model. This suggests various levels of darkness map to the same brightness. Although such ablation helps our model on LOL and MIT-Adobe-FiveK datasets, it does not on our test-set. We do not rely on these small size datasets (LOL with 15 images and MIT-Adobe-FiveK with 500 images) to choose our proposed model, but we consider the overall performance across all three datasets with ours being the most representative and diverse set (4k images).

{\bf w/o noise}.
Generating low light images by ablating noise augmentation means ignoring the noise caused by cameras on real images taken in the dark. Such ablation only helps our model on LOL test-set consisting of 15 images. It does not help our model on the MIT-Adobe-FiveK and our test-sets, suggesting it is not safe to ignore the noise augmentation step.

{\bf w/o CRF}.
Generating low light images by ablating camera response function in the procedure means we ignore the precise measurement of scene radiance. We do not know the true CRF for an image, but we use a random CRF from the database of real-world camera response functions (DoRF)~\cite{crf} which Grossberg et al. show these responses occupy a small part of the theoretical space of all possible responses. Thus we use their empirical model of response (EMoR) to accurately interpolate the camera response function and linearize a given image.
Such ablation helps our model perform better on MIT-Adobe-FiveK and our test-sets, but it does not help the model to generalize well to the LOL test-set.

{\bf w/o constant $k$}.
Passing low light images to the model without multiplying them by constant $k$ drops the model performance on the LOL and MIT-Adobe-FiveK test-sets, affecting the model's generalization capability.

{\bf w/o LAB color space}.
Passing low light images ablating conversion to LAB color space drops the model performance on LOL and MIT-Adobe-FiveK test-sets,  affecting the model's generalization capability.

{\bf w/o adversary}. 
Supervising the model without an adversarial loss affects model performance as it is not able to generalize well to the LOL and MIT-Adobe-FiveK test-sets.

Our \textbf{proposed model} uses a combination of 4 losses, takes input images in LAB color space multiplied by constant $k$, is trained on data generated by our pipeline (scaling the original input by $1+\epsilon$, passing through an inverse CRF to linearize, augmenting with noise) and performs best across all three test-sets, our test-set generated by our pipeline as well as the standard LOL and MIT-Adobe-FiveK test-sets.

\begin{table}{}
\begin{center}
\caption{Ablation studies in terms of PSNR (in dB), SSIM, and LPIPS on various test-sets.}
\label{table:ablations}
\resizebox{\textwidth}{!}{\begin{tabular}{| l | S | S | S | S | S | S | S | S | S |}

\cline{2-10}
\multicolumn{1}{c|}{}& \multicolumn{3}{c|}{Ours} & \multicolumn{3}{c|}{LOL} & \multicolumn{3}{c|}{MIT-Adobe-FiveK} \\
\hline
Ablations & {PSNR}$\uparrow$ & {SSIM}$\uparrow$ & {LPIPS}$\downarrow$  & {PSNR}$\uparrow$ & {SSIM}$\uparrow$ & {LPIPS}$\downarrow$ & {PSNR}$\uparrow$ & {SSIM}$\uparrow$ & {LPIPS}$\downarrow$\\
\hline
w/o $1+\epsilon$ & 28.023 & 0.840 & 0.020 & 23.050 & 0.821 & 0.058 & 21.485 & 0.748 & 0.116\\
w/o noise & 27.827 & 0.847 & 0.029 & 23.104 & 0.750 & 0.080 & 19.726 & 0.714 & 0.151\\
w/o CRF & 31.586 & 0.900 & 0.009 & 20.188 & 0.789 & 0.076 & 23.721 & 0.789 & 0.090\\
w/o constant $k$ & 28.161 & 0.834 & 0.019 & 12.088 & 0.696 & 0.085 & 10.516 & 0.603 & 0.119\\
w/o LAB & 28.589 & 0.840 & 0.025 & 15.396 & 0.300 & 0.306 & 12.481 & 0.608 & 0.217\\
w/o adversary & 28.318 & 0.840 & 0.021 & 19.232 & 0.777 & 0.083 & 17.100 & 0.693 & 0.166\\
Proposed & 28.140 & 0.835 & 0.021 & 21.485 & 0.820 & 0.057 & 20.344 & 0.754 & 0.109\\
\hline
\end{tabular}}
\end{center}
\end{table}

\begin{figure}
\scriptsize
\stackunder[1pt]{\includegraphics[width=0.30\textwidth]{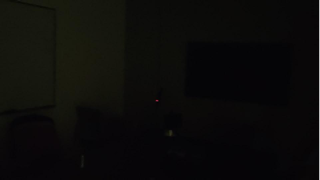}}{(a) Input}
\stackunder[1pt]{\includegraphics[width=0.30\textwidth]{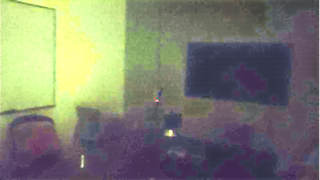}}{(b) Output}
\stackunder[1pt]{\includegraphics[width=0.18\textwidth]{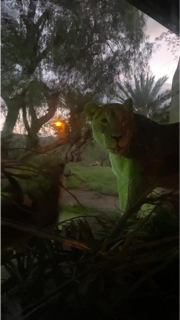}}{(c) Input }
\stackunder[1pt]{\includegraphics[width=0.18\textwidth]{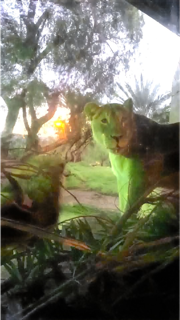}}{(d) Output}
\caption{\label{fig:limitation} Failure examples on the LLIV-Phone-ImgT test-set.}
\end{figure}

Some failure cases of the proposed method are shown in Fig.~\ref{fig:limitation}. As can be seen in these examples, the proposed method may fail on enhancing images taken in extremely dark environments. It introduces some artifacts, color shifts and tendency towards greenish colors when met with these challenging examples. For future work, we hope to strengthen our proposed solution to provide visually acceptable results even in such extremely low light conditions.

\subsection{Experimental Findings}
Tab.~\ref{table:quan_res} summarizes our results. Our model outperforms the standard competitive methods on standard test-sets using standard metrics.  Our model is best at LOL on all standard metrics by a significant margin. On MIT-Adobe-FiveK, our model is best in one metric, and in the top three
for the two others. Note our model has seen no image from either set; most models are trained on images from one or the other. On LLIV-Phone-ImgT, our method ranks in the middle of competitive methods. But this dataset is evaluated with non-reference metrics NIQE and SPAQ.  We place less weight on ranking by such metrics, because they do not have access to ground truth when assessing a result.  

Tab.~\ref{table:ablations} shows that on occasion ablations improve performance on a dataset. But when an ablation improves performance on one dataset, it hurts performance on another. Our method peforms very well across all datasets and metrics, and demonstrates strong generalization capability, and outperforms current state of the art methods. We believe this is because the very great volume of data obtainable by simulating a pipeline outweighs the effects of possible errors in simulation.  



\section{Conclusions}
In this paper, we study the problem of making brighter images from dark images found in the wild. The images are dark because they are taken in dim environments. They suffer from color shifts caused by quantization and from sensor noise. We don't know the true camera reponse function for such images and they are not RAW. We use a supervised learning method, relying on a straightforward simulation of an imaging pipeline to generate usable dataset for training and testing. On a number of standard datasets, our approach outperforms the state of the art quantitatively. Qualitative comparisons suggest strong improvements in reconstruction accuracy.
\bibliographystyle{splncs04}
\bibliography{egbib}

\end{document}